\documentclass[transmag]{IEEEtran}
\pdfoutput=1 
\usepackage{amsmath,amssymb,amsfonts}
\usepackage{algorithmic}
\usepackage{graphicx}
\usepackage{textcomp}
\usepackage{textcomp}
\usepackage{latexsym}
\usepackage{graphicx}
\usepackage{array}
\usepackage{amsfonts,amssymb,amsmath}
\usepackage{hyperref}
\usepackage{balance}
\usepackage[center]{caption}
\usepackage{setspace}

\def\BibTeX{{\rm B\kern-.05em{\sc i\kern-.025em b}\kern-.08em T\kern-.1667em\lower.7ex\hbox{E}\kern-.125emX}}

\newcolumntype{P}[1]{>{\centering\arraybackslash}m{#1}}

\begin{document}

\title{A Survey on the Use of AI and ML for Fighting the COVID-19 Pandemic}

\author{Muhammad Nazrul Islam, Toki Tahmid Inan, Suzzana Rafi, Syeda Sabrina Akter, Iqbal H. Sarker, A. K. M. Najmul Islam
\thanks{This work was submitted on Date of submission}
\thanks{Muhammad Nazrul Islam is with the Department of Computer Science and Engineering, Military Institute of Science and Technology, Dhaka, Bangladesh. (email: nazrul@cse.mist.ac.bd) }
\thanks{Toki Tahmid Inan is  with the the Department of Computer Science, George Mason University, VA 22031, USA. }

\thanks{Suzzana Rafi is with the Department of Computer Science and Engineering, Bangladesh University of Engineering and Technology, Dhaka, Bangladesh.}
\thanks{Syeda Sabrina Akter is is with the Northern University Bangladesh, Dhaka, Bangladesh. }

\thanks{ Iqbal H. Sarker is with the Department of Computer Science and Engineering, Chittagong University of Engineering and Technology, Chittagong, Bangladesh.}  

\thanks{ A. K. M. Najmul Islam is with the Department of Future Technologies,
University of Turku, Finland and LUT school of Engineering, LUT University, Finland.}
}

\maketitle
\begin{abstract}Artificial intelligence (AI) and machine learning (ML) have made a paradigm shift in health care which, eventually can be used for decision support and forecasting by exploring the medical data. Recent studies showed that AI and ML can be used to fight against the COVID-19 pandemic. Therefore, the objective of this review study is to summarize the recent AI and ML based studies that have focused to fight against COVID-19 pandemic. From an initial set of 634 articles, a total of 35 articles were finally selected through an extensive inclusion-exclusion process. In our review, we have explored the objectives/aims of the existing studies (i.e., the role of AI/ML in fighting COVID-19 pandemic); context of the study (i.e., study focused to a specific country-context or with a global perspective); type and volume of dataset; methodology, algorithms or techniques adopted in the prediction or diagnosis processes; and mapping the algorithms/techniques with the data type highlighting their prediction/classification accuracy. We particularly focused on the uses of AI/ML in analyzing the pandemic data in order to depict the most recent progress of AI for fighting against COVID-19 and pointed out the potential scope of further research.
\end{abstract}

\small\par \textbf{\textit{Impact Statement}}
 — The application of Artificial intelligence (AI) has created a paradigm shift in healthcare. From disease detection  to pandemic forecasting, ubiquitous usage of AI has been proven promising in healthcare.  The recent novel corona virus: COVID 19 has thrown a challenge to the researchers and the health professionals. Therefore, the researchers are seeking help from AI to fight the latest pandemic COVID-19. In this review article we aim to explore and analyze the research work that focus on the usage and application of AI and machine learning to fight against COVID-19.


\section{INTRODUCTION}
\IEEEPARstart{T}{he} novel and contagious viral pneumonia, COVID-19 (Coronavirus disease-2019) has affected more than 8.2 million people and caused death of more than 440,000 thousand people worldwide. WHO declared it as a global pandemic\cite{who} and suggested that early detection, isolation and prompt treatment can be useful to slow down the COVID-19 outbreak \cite{who2}. Therefore, various bodies have committed themselves to conduct research focusing on COVID-19 to support global response.

With new discoveries being announced at a breathtaking pace, artificial intelligence (AI) has re-emerged into scientific consciousness. AI is a branch of computer science that can be used to build intelligent systems. It is often instantiated as a software program \cite{ligeza}. Recent application of AI in diagnosing disease s has broadened the frontier of AI which once was a humans’ expertise. Medicine and the health care systems are among the most promising areas of application of AI which can be traced back to as early as mid-twentieth century\cite{ai}. Researchers proposed and successfully developed several decision support systems\cite{jamia}. The rule-based system gained success in the late ‘70s\cite{aim} and had been useful in detecting disease\cite{abdominal}, interpreting ECG images\cite{ecg}, choosing appropriate treatment and generate hypothesis by the physicians\cite{inter}. Unlike this first-generation knowledge-based AI system, which relies upon the prior medical knowledge of experts and the formulation based rules, modern AI leverages machine learning algorithms to find pattern and associations in data \cite{circ} \cite{sarker2019effectiveness} \cite{sarker2018mining}. The recent renaissance in AI can be attributed to the successful application of deep learning to a great extent- training an artificial neural network with a large number of labelled datasets. A modern deep learning network usually contains hundreds of hidden layers\cite{nature}. The recent resurgence of AI has been fueling a question of whether AI-doctors will replace human physicians shortly. Yet to be confirmed, researchers believe that AI driven intelligent systems can significantly help human physicians in making better and quick decisions and even sometimes remove the necessity of human decision i.e. radiology\cite{vasc}. The increasing data in health care resulting from increased use of digital technology and the advancement of big data analytics can be attributed to the recent success of AI in healthcare\cite{vasc}. Although AI research in healthcare is emerging, most of the research is concentrated on three diseases: cancer, neurology and cardiology. Guided by evidence, a strong AI can reveal the insight of the medical data which eventually can be used for decision support and forecasting\cite{jama}, \cite{omics}, \cite{ccr}.
 
As AI has been proved useful in healthcare, researchers suggest that it may also be helpful in fighting against COVID-19. From forecasting of a pandemic to designing anti-viral-replication molecules, AI has made a paradigm shift in health care. Recent research on COVID-19 using AI suggests that AI can be helpful in detecting COVID-19 infection, detecting infected population, predicting the next outbreak, finding the attack pattern and even finding a cure \cite{artificial}, \cite{Wang} , \cite{con}. The objective of this review is to explore the existing AI-based research that has been conducted to fight against COVID-19 pandemic. 
The organization of the remaining section is as follows. The methodology to conduct this review study is discussed in section 2. The review data analysis and findings are discussed in detail in section 3. Section 4 presents the main findings and the potential scopes of future research to fight against the COVID-19. Finally, the concluding remark, research limitation and future work are presented in section 5.

\section{METHODOLOGY}

In this research, a systematic literature review procedure \cite{acq} was adopted to attain the research aim. For selecting the primary articles, the major databases such as IEEE Xplore, Springer Link, ACM digital library, Science Direct, and Google Scholar were searched for related articles. The search strings used to find the literature were “Machine learning and COVID-19”, “Machine learning and Coronavirus”, “Artificial Intelligence and COVID-19”, “Artificial Intelligence and Coronavirus”, “Artificial Intelligence and lockdown and pandemic and COVID-19”, “Machine learning and lockdown and pandemic and COVID-19”, and “Coronavirus prediction and outbreak prediction and machine learning and artificial intelligence”. These strings were applied for all the above-mentioned databases as well as Google search engine.

\begin{figure}[htp]
\begin {center}
\includegraphics[width=3.3in]{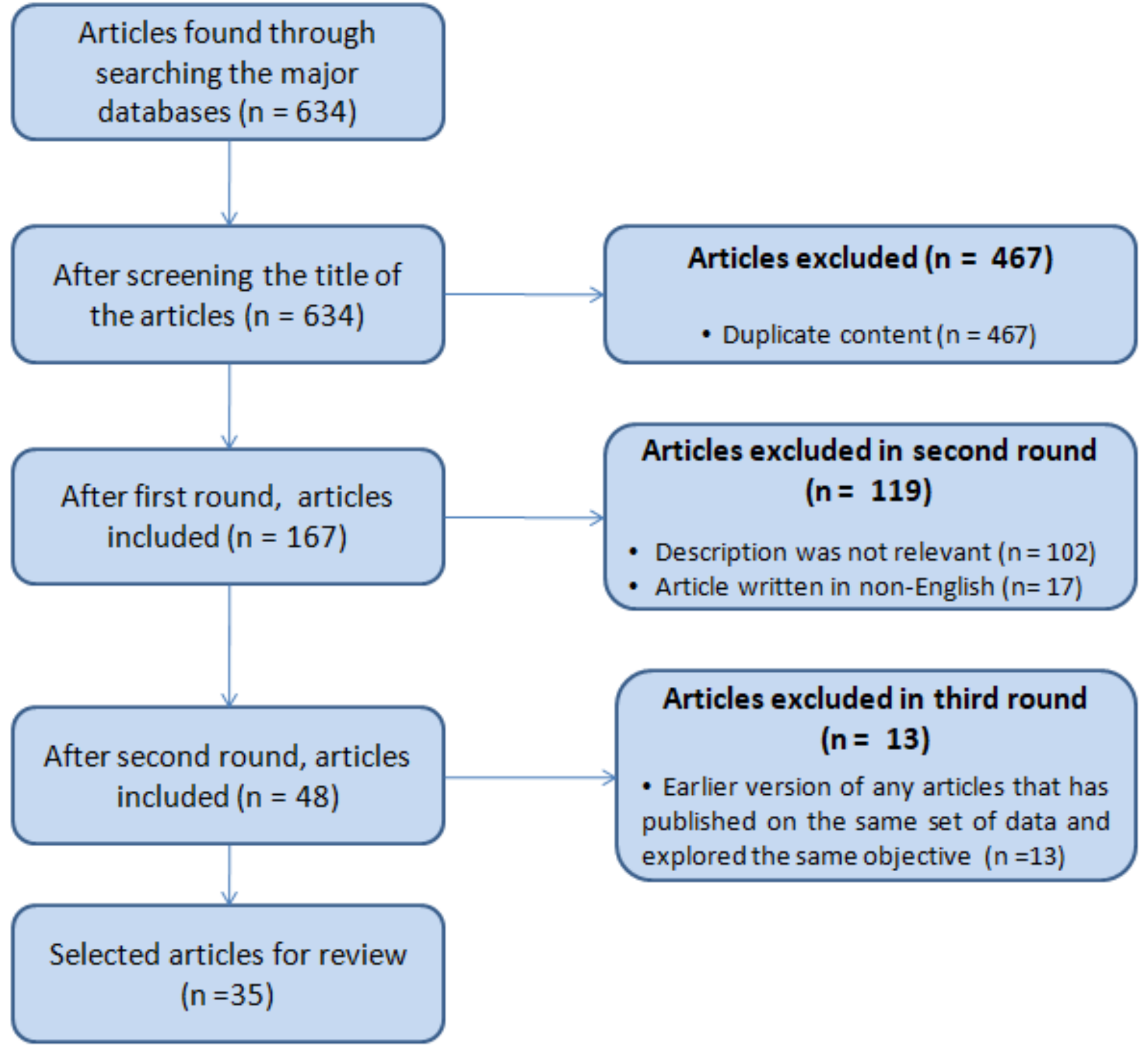}
\caption{Article inclusion and exclusion process flowchart}
\label{fig:f1}
\end {center}
\end{figure}

The search returned more than 634 articles. The exclusion criteria include (a) duplicate articles that are found through several scholarly databases; (b) articles that are not focused on our research objectives; (c) the article was written other than English; and (d)  the earlier version of any article that has been published on the same set of data and explored the same objective. After applying this inclusion-exclusion process, we finally selected 35 articles that include original research, review articles, and short articles including perspective, commentary, and letter to the editor. The Prisma flowchart in Figure \ref{fig:f1}  shows the article selection process in different phases following the exclusion-inclusion criteria.

The selected articles were reviewed systematically to extract data primarily related to the article type, publication time, research objectives, study context, study outcomes/findings, methodology/algorithm/techniques used, dataset used, and study subject. Finally, the extracted data were synthesized and analyzed to summarize the existing research on COVID-19 pandemic using AI and to identify the potential scopes for future research.

\section{DATA ANALYSIS AND FINDINGS}

\subsection{Type of Publications}

To date (12th June, 2020) out of 35 articles, 27 (77\%) were published as original research. Among others (23\% articles), three articles were the review articles, two were editorial and the remaining three were published as research perspective (short conceptual article). Again, 21 (60.0\%) articles were published in academic journals, while 14 (40\%) articles were archived as pre-print. Among the pre-print, 92.85\% (13) were original research. All the selected articles were published or archived in the online databases during January-May 2020.  

\subsection{Research Purposes and Objectives}

We synthesise the existing research in terms of their purposes and aims to explore the contribution of AI to fight against the COVID-19 pandemic. A summary of the synthesized data is presented in Table \ref{tab:table_purposes} to show the research scopes and purposes of the original research. Most of the articles (n =16, 48\%) were published focusing to detect the COVID-19 infected patients using different AI-based algorithms that include, for example, the Convoluted Neural Network (CNN) model, Support Vector Machine (SVM), generative adversarial network (GAN), and the transfer learning. The chest X-ray images, CT images, mobile sensors data, and  COVID-19 symptoms were used to predict/detect the COVID-19 patients. These researches aimed to identify, screen and detect the COVID-19 patients; and also to predict, differentiate, or classify the patients into COVID-19 infection, no infection, and other viral or bacterial infection. For example, Wang et al. \cite{covidnet} proposed a CNN based prediction system named COVID-Net that identifies non-COVID-19 infected, COVID-19 infected, and no infected patients using chest X-ray images. The proposed model was pre-trained on ImageNet (open source) dataset and then trained on the COVIDx (author- created) dataset that includes 13,800 chest X-ray images of 13,725 patients that includes 183 images from 121 COVID-19 positive patients, 8066 images are of healthy patients and 5538 images are of non-COVID19 patients.

\begin{table*}[]
\begin{singlespace}
\small
\centering
\caption{A summary of the literature based on their purposes}
\label{tab:table_purposes}

    \begin{tabular}{{|P{3cm}|P{10cm}|P{1.3cm}| P{1.3cm}|}}
    \hline
    Purposes & Brief Description & Reference & Frequency \\
    \hline
    
      
          & Identify the infected individual quicker	 &  \cite{con} &  \\ 
       \cline{2-3}

       & Screen coronavirus diseases using deep learning & \cite{a26} & \\
       
      \cline{2-3}

        & Identify the coronavirus patients & \cite{a27} & \\
       
      \cline{2-3}
      
        &   Developing a CNN based Algorithm to detect COVID-19 from CT images & \cite{Wang} & \\
       
      \cline{2-3}

        & Detect COVID-19 with the help of AI and smartphone sensors & \cite{a28} & \\
       
      \cline{2-3}
      
      & Use an anomaly model based on deep learning network to make the screening process faster for Covid-19 detection from X-ray images & \cite{covidnet} & \\
       
      \cline{2-3}
      
      & Detect COVID-19 from X-ray images using transfer learning with CNN & \cite{a29} & \\
       
      \cline{2-3}
      
      & Detect COVID-19 from X-ray images using deep CNN model & \cite{a30} & \\
       
      \cline{2-3}
      
        &Propose an algorithm to detect COVID-19 from CT images using deep CNN model and SVM classifier & \cite{a31} & \\
       
      \cline{2-3}
      
        & Develop a deep learning model CoroNet using the Xception CNN to detect Covid-19 from Xray images & \cite{a32} & \\
       
      \cline{2-3}
      
        &Build a framework which uses smartphone sensors to detect Covid-19 & \cite{a28} & \\
       
     \cline{2-3}

       Diseases detection  & Classify patients in to non-COVID 19 infection, COVID-19 infection, and no infection from X-ray images using deep CNN model & \cite{covidnet} &16 \\
       
      \cline{2-3}

        & Compare the performance of seven DL model to find the best model for Covid-19 detection & \cite{a33} & \\
       
      \cline{2-3}

        & Develop and evaluate the performance of an AI model to detect Covid-19 and also evaluate the performance of radiologist to detect the disease by using and without AI support & \cite{a34} & \\
       
      \cline{2-3}

         & Detect the  Covid-19 by identifying the characteristics from chest X-ray using a deep learning model(CAD4COVID-XRay)& \cite{a35}& \\
       
      \cline{2-3}

         & Detect COVID-19 from X-ray images using Generative Adversarial Network (GAN) and Deep learning transfer & \cite{a36} & \\

      \cline{1-4}

         & Diagnosis the identified patients to classify (in to patients' categories) and tracking the progress Covid-19 patients & \cite{a37} & \\
       
      \cline{2-3}

         & Distinguish COVID-19 from pneumonia using Deep learning & \cite{a38} & \\
       
      \cline{2-3}
       
       &  Efficiently diagnosis COVID-19 using X-ray images through deep CNN models & \cite{a39} & \\
             \cline{2-3}
      
         & Developing a tool to predict survival and death for severe COVID-19 patients & \cite{a40} & \\
       
      \cline{2-3}

         Diseases diagnosis & Diagnosis Covid-19 positive case faster using both non-image and image clinical data   & \cite{a41} & 7  \\
       
      \cline{2-3}

          & Develop a system to identify patients who would develop more severe illness among the patients with mild cases of COVID-19  & \cite{a42} & \\
       
      \cline{2-3}

         & Develop a system to improve the diagnostic performance from Posterior-Anterior (PA) X-ray images of lungs with COVID-19 cases  & \cite{a43} & \\
       
      \cline{1-4}

          & Forecast of the COVID-19 to estimate size, lengths and ending time of COVID-19 across China  & \cite{artificial} & \\
       
      \cline{2-3}

       Epidemic forecasting & Predict the trend of the infection for the next 80 days using deep learning  & \cite{a45} & 2\\
       
      \cline{2-3}

          & To predict the progress of the epidemic (epidemic sizes and peaks) & \cite{a45} & \\
       
      \cline{1-4}

         Sustainable development & Analyze the correlation among environmental factors and confirmed cases of COVID-19 & \cite{a46} & 1 \\
       
      \cline{1-4}

          & Compare the prediction performance of the proposed algorithm with the existing methods & \cite{a31} & \\
       
      \cline{2-3}

       & Compare seven different DL model to find out the best model for disease detection   & \cite{a33} & \\
       
      \cline{2-3}

        Performance comparison  & Compare the performance of radiologist in distinguishing Covid-19 from other pneumonia with and without AI assistance  & \cite{a34} & 4\\
       
      \cline{2-3}

          & Compare the performance of a DL model with six other radiologists   & \cite{a35} & \\
       
      \hline

             Patient management  & To improve management of Covid-19 ICU patients   & \cite{a47} &  1\\
       
      \cline{1-4}
      
    \end{tabular}
\end{singlespace}
\end{table*}

\begin{figure}[hbp]
\begin {center}
\includegraphics[width=3.4in]{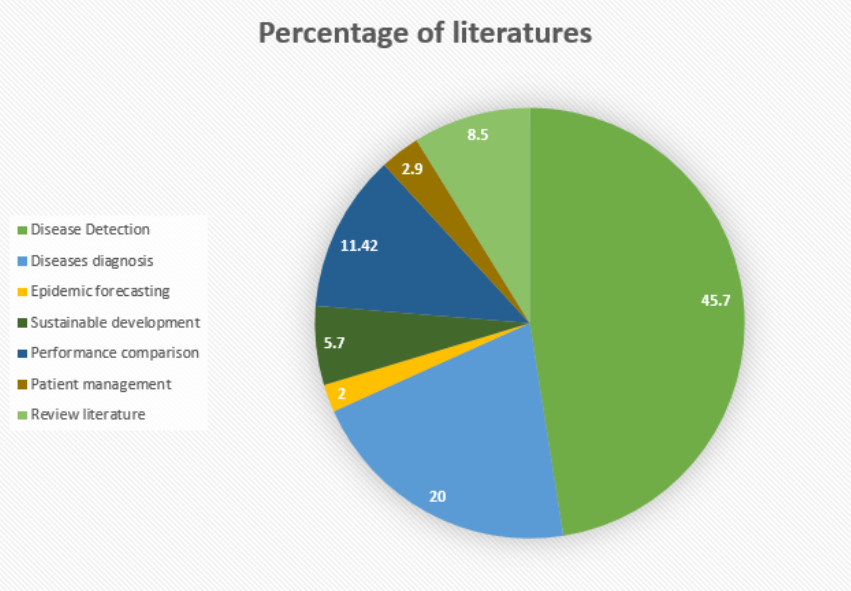}
\caption{Percentage of papers based on objectives}
\label{fig:f2}
\end {center}
\end{figure}

A total of seven (20\%) articles focused on diagnosing COVID-19 patients through AI (Figure \ref{fig:f2}). In these articles, the AI was used to diagnose the identified COVID-19 patients to classify in to patients’ categories (severe, mild) and tracking their progress\cite{a27}, distinguish COVID-19 from pneumonia \cite{a38}, efficiently diagnose COVID-19 using X-ray images\cite{a39}, predict survival and death for severe COVID-19 patients\cite{a40}, identify patients who would develop more severe illness \cite{a42}, and to estimate uncertainty in Deep Learning solutions to improve the diagnostic performance from Posterior-Anterior (PA) X-ray images of lungs with COVID-19 cases \cite{a43}.

Two (6\%) articles aimed to forecast the COVID-19 epidemic to estimate the progress of the epidemic in terms of its size, lengths, peaks, and ending time as well as predict the development trend of the epidemic for the next certain time period in a specific country or geographical region \cite{artificial}\cite{a45}. We found only one study \cite{a46} focused the sustainable development that analyzes confirmed cases of COVID-19 through a binary classification using AI and regression analysis and explores the correlation among confirmed cases of  COVID-19 in four countries (China, Italy, South Korea, Japan) and environmental factors (low, high \& average temperature, humidity, wind flow). An article \cite{a31} compared the prediction performance of the proposed algorithm with the existing VGG-16, GoogleNet, and ResNet50 method using two different subsets of data, while in \cite{a47}, an AI-based model is developed based on the existing studies regarding AI in ICU and respiratory diseases to improve the (COVID-19) ICU patient management.

The remaining (26\%) articles include review articles, editorial, perception, commentary, and short communication. The review articles are summaries of the existing research with the aim to highlight the contributions and constraints of AI \cite{a48}, and to identify a roadmap of AI applications to fight against COVID-19 pandemic \cite{a49}. Another review analyzes the AI-based techniques used in the CT and X-ray based medical imaging to fight against COVID-19 pandemic \cite{acq}. One of the two editorials highlighted how AI-based solutions may assist to fight against the pandemic by forecasting the pandemic to design anti-viral replication molecules, but with the supervision of humans \cite{a50}. In another editorial, a workflow is presented to highlight the processes and applications of AI to fight the COVID-19 pandemic \cite{a51}. The perspective articles highlight firstly the needs of AI and the ways of data sharing (via smart city networks) for better monitoring and management of urban health on the COVID-19 outbreak \cite{a52}; secondly, the importance of active learning-based AI tools for coronavirus outbreak \cite{a53}; and finally, suggested how AI and Blockchain can be used to help the community during the COVID-19 pandemic with equipment and donations \cite{a54}. By using a private blockchain network to make donations for the pandemic, there would be no alterations and the donations would go to their destinations. A summary of the synthesized data is briefly presented in Table \ref{tab:table_purposes_OTHER} to present the research scopes and purposes of the other types of research.

\begin{table*}[t]
\small
\centering
\caption{Scopes of other types of research}
\label{tab:table_purposes_OTHER}

    \begin{tabular}{{|P{3cm}|P{8cm}|P{1.3cm}| P{1.3cm}|}}
    \hline
    Purposes & Brief Description & Reference & Frequency \\
    \hline
    
      
          & Review the related work to highlight the contributions and constraints of AI in fighting the Covid-19 pandemic	 & \cite{a48} &  \\ 
       \cline{2-3}
        Review Literature & Review related work to identify a roadmap of AI applications to fight against the pandemic & \cite{a49} & 3 \\
       
      \cline{2-3}
      
      & Review the AI based techniques used in the CT and X-ray based medical imaging data acquisition, segmentation and diagnosis to fight against COVID-19 pandemic & \cite{acq} & \\
      
      \cline{1-4}

         Editorial & Highlight how AI based solutions may assist to fight against the COVID-19 pandemic & \cite{a50} & 2  \\
       
      \cline{2-3}

         & The editorial constitutes existing works, current efforts and potential work ideas to fight against Covid-19 using AI, ML algorithms, deep learning, neural networks. & \cite{a51} &  \\
       
      \cline{1-4}

          & Highlighted the needs of AI and the ways of data sharing via smart city networks for a better monitoring and management of urban health on the COVID-19 outbreak & \cite{a52} &  \\
       
      \cline{2-3}

         Perspective & Discussed the importance of active learning based AI tools for coronavirus outbreak. Tools that use cross population training/testing methods and multitudinal and multimodal data. & \cite{a53} & 3 \\
          
          \cline{2-3}
          
          &  Introduced AI and Blockchain and suggested how they can be used to effectively help the community with equipment and donations.Introduced AI and Blockchain and suggested how they can be used to effectively help the community with equipment and donations. & \cite{a54}&  \\
       
      \cline{1-4}

    \end{tabular}

\end{table*}

\subsection{Context of Study}

Some of the articles focused their research on specific countries while others conducted research with a global perspective. A total of 10 articles (29\%) focused on specific country as shown in Table \ref{tab:table_details_contexual}. One of these articles considered confirmed cases from 34 provinces of China for their research on a forecasting system of those areas \cite{artificial}. Another article focused on 42 provinces in Japan, China, South Korea and Italy for environmental parameters, weather trends and confirmed cases to measure correlations and also build a classification model \cite{a46}. In one study CT scan of lungs from patients of both USA and China \cite{a27} were used. In another study, CT scan of lungs only from China \cite{Wang} were used for training and testing automated AI-based tools for diagnosis and tracking. Epidemiological data of three provinces of China (Hubei, Guangdong and Zhejiang), SARS 2003 epidemic data of all over China were collected and the prediction was made for the whole China \cite{a45}. CT images of patients of Italy were used as well in another study \cite{a31}. Data was collected from only Wuhan, China in one of the articles \cite{a40} and Wenzhou city of Zhejiang province in another article \cite{a42}. As we see most articles concentrated on data from China as it is the original epicentre of the pandemic.

\begin{figure}[htp]
\begin {center}
\includegraphics[width=3.5in]{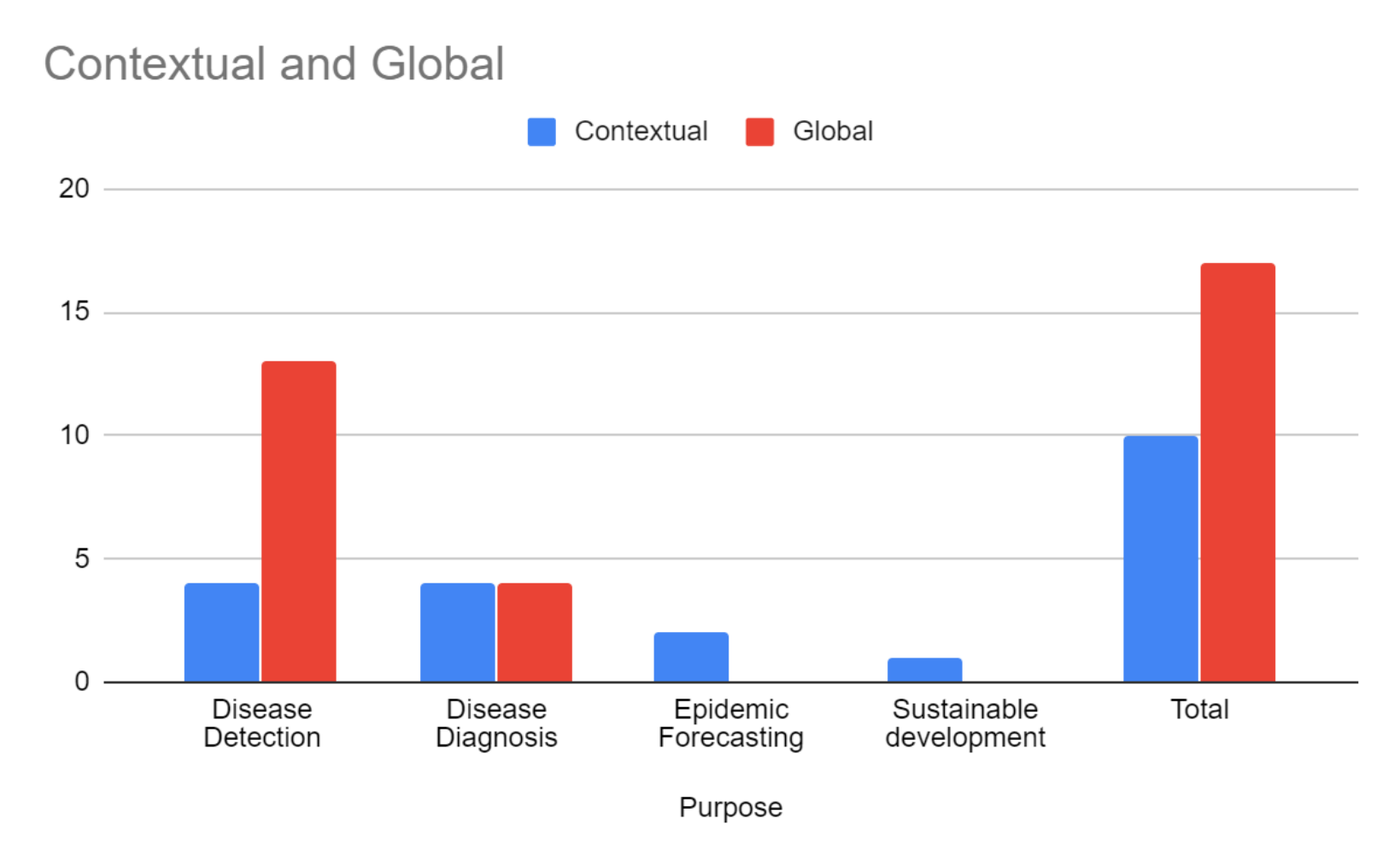}
\caption{A brief on the context of data used by the literature}
\label{fig:f3}
\end {center}
\end{figure}

Contextual articles focused mainly on epidemic forecasting and sustainable development. Most of the disease detection related articles and all of the recommendation type articles used global perspectives as well as public datasets and are not context-sensitive (Figure~\ref{fig:f3}). Thus, it can be said that the disease detection techniques mentioned in the articles are mostly not context-dependent; and for the epidemic forecasting purpose researchers need contextual data.
There were two cross-country studies - one of them \cite{a46} focused on finding correlation among multiple countries’ COVID-19 cases and the other study \cite{a27} focused on two different country cases to enhance the performance of their disease detection tool. Table \ref{tab:table_details_contexual} briefly shows the details on the data used in these contextual studies.

\begin{table*}[htp]

\caption{A brief details on Data of the contextual literatures}
\label{tab:table_details_contexual}
\small
\centering

    \begin{tabular}{|p{1.6cm}|P{1.5cm}|P{4.5cm}| P{4.5cm}| P{4cm}| }
    \hline
    Literature &	Objective	&Data Source & 	Data Volume	& Data Type \\
    \hline
    
    Hu et al. \cite{artificial} & 
    Epidemic forecasting &
 	WHO and local Chinese news media collected Data & 
 	15,384 and 36,602 cases Clinically confirmed and lab confirmed cases respectively & 
 	Time series data (Non -Image) \\
 	
 	\hline

 	Pirouz et al. \cite{a46} & 
    Sustainable development & 
 	Data from 42 province of China, Japan, Italy and South Korea & 	        	-	&
 	Environmental, geographical and demographical data from 28 January 2020 to 26 February 2020(Non -Image)
 	\\
   \hline

   Gozes et al.\cite{a27} & 
   Diseases diagnosis & 
   \begin{enumerate}
       \item Development Dataset Source: Chainz
       \item Testing Dataset Source: Hospital in Wenzhou, China, Chainz, El-Camino Hospital (CA), LIDC
       \item Lung segmentation Development Sources: El-Camino Hospital (CA)
sources were used
   \end{enumerate}

 	& 157 patients 
 	& CT scan images of lungs(Image)
 	\\
 	\hline

 	Wang et al \cite{a55} & 
Disease detection & 
 	China & 
 	453 images from 99 patients	 & 
 	CT images of chest (Image)
 	\\
 	\hline

 	Yang et
al. \cite{a45} & 
 	Epidemic forecasting & 
 	Covid-19 outbreak data reported by the National Health Commission of China(Wuhan, Hubei province, Guangdong province, Zhejiang province) , Migration data was retrieved from a web based program, 2003 SARS epidemic data was retrieved from an archived news-site (SOHU) & 
 	  -	& Non -Image
 	  
 	  \\
 	\hline

 	Umut et al. \cite{a31} & 
 	Disease Detection & 
 	Societa Italiana di Radiologia Medica e Interventistica (Itali) & 
 	150 CT images   & 	Time series data(Non -Image)
 	
 		  \\
 	\hline

 	Li et al.\cite{a40} & 
 	Diseases diagnosis & 
 	Wuhan (China) clinical Data	& 3129 cases of COVID-19 patients & 
 	Time series(Non -Image)
 	
 			  \\
 	\hline
 	
 	Xiangao et al. \cite{a42} & 
 	Disease Detection & 
 	Clinical data from Wenzhou, Zhejiang, China. & 
 	53 hospitalized patients'
 	& Medical data (Non -Image) 
 			  \\
 	\hline

 	Mei et al.\cite{a41} & 
Disease Diagnosis & 
	Chest CT studies and clinical data from China & 	905 patients  & 	Chest CT images
And clinical data (Non-image)

			  \\
 	\hline
 	
 	Harison et al. \cite{a34} & 
Disease Detection & 
	Chest Xray from Hunan province, China &  
	512 patients 
	&	Chest X-ray (Image)
			  \\
 	\hline
    

    \end{tabular}

\end{table*}

\subsection{Exploration of the Used Data Type}

\begin{figure*}[t]
\begin {center}
\includegraphics[width=5in]{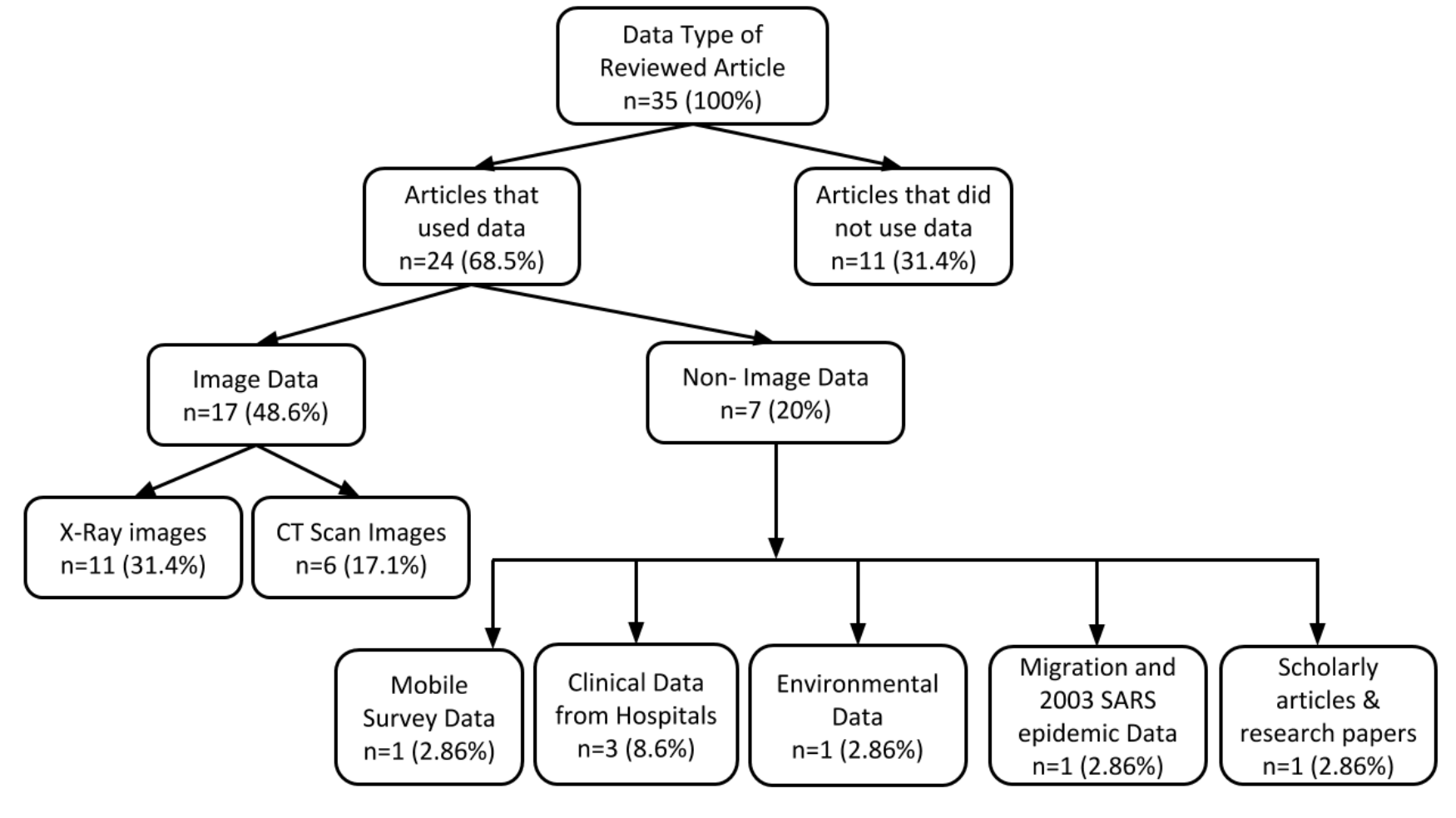}
\caption{Analyzing the Data used in the reviewed articles}
\label{fig:f4}
\end {center}
\end{figure*}

24 studies (68.5\%) had used different types of data, ranging from text to image, to corroborate their findings as shown in Figure \ref{fig:f4}.
The higher percentage (63\%) with 22 original studies prove the dependency of AI and ML based systems on proper data assessment.

Seventeen (40\%) of the articles used image data in the form of X-Ray images and CT images of the chest. These images are mostly used in disease detection and diagnosis. Seven studies (23.3\%) used public datasets among which four studies used the COVID-19 dataset from GitHub repository created by Dr. Joseph Cohen, a postdoctoral fellow at the University of Montreal. X-Ray images for other lung disease patients, such as pneumonia, were collected from GitHub repository, Kaggle repository and Open-I repository \cite{a56}. Two of the other studies collected data from various hospitals from China and the USA (Table \ref{tab:table_algorithm}). One study collected data from Societa Italiana di Radiologia Medica e Interventistica \cite{a28}, a hospital from Italy. 

The other seven articles (20\%) used non-image data, predominantly in the form of text and numbers with the purpose of disease detection, epidemic forecasting, sustainable development, introducing advanced concepts, and disease diagnosis and progression (Table \ref{tab:table_purposes}). Thousands of data points were collected through a mobile survey for a study \cite{con} that included information related to location, age, gender, race, travel and close contact to any affected person. Furthermore, the study collected health data related to COVID-19 symptoms during a period of 14 days. Clinical data including information on baseline characteristics, medical history, COVID-19 diagnosis from hospitals in Wuhan and other provinces in China were collected for disease prediction and progression purposes. The study \cite{a46} used environmental and urban data accumulated from 42 different provinces in China, Japan, South Korea, Italy to analyze the correlation among environmental factors (low, high and average temperature, humidity, wind flow) and confirmed cases of COVID-19. The study here \cite{a45} used migration data and 2003 SARS epidemic for epidemic forecasting of COVID-19. Finally, the study\cite{a51} used various research papers and scholarly articles for the purpose of proposing potential ideas to combat COVID-19 using AI.

We observed that nine (25.8\%) studies used data that were collected from different provinces in China making China as the major source of initial COVID-19 related Data.

\subsection{Exploring the AI Techniques}

Most of the research papers (n=29, 83\%) aim to use AI to do some kind of classification (COVID-19 detection, differentiate COVID-19 from other respiratory diseases) forecasting, and prediction (Table \ref{tab:table_purposes}). The cost and time associated with the gold standard of testing COVID-19: PCR - takes up to two to three days to get the results, drive researchers to find an easier, cheaper and faster way to detect COVID-19 using computational technique. Therefore, our study found most of the research work (65\%) aimed to detect and diagnose COVID-19. Table \ref{tab:table_algorithm} briefly presents the objective, scope and the results of using different AI algorithms. 

\begin{figure}[htp]
\begin {center}
\includegraphics[width=3.6in]{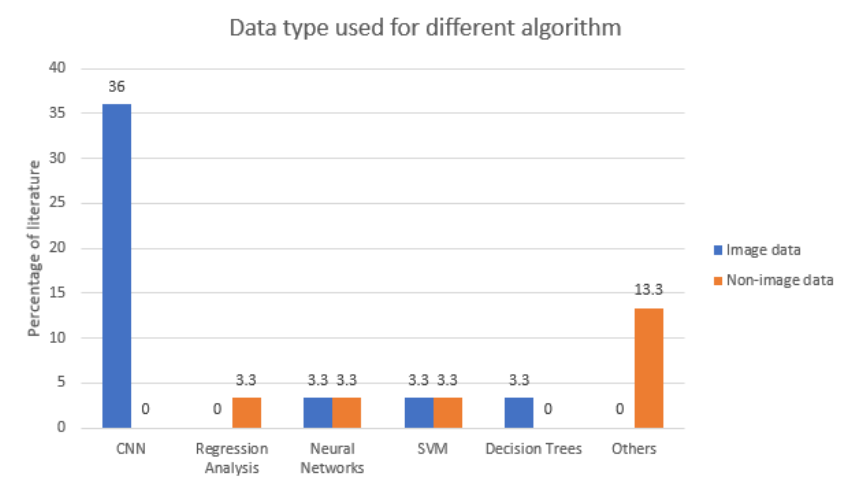}
\caption{The percentage of literature used different data type for different algorithm}
\label{fig:f5}
\end {center}
\end{figure}

\begin{figure}[htp]
\begin {center}
\includegraphics[width=3.5in]{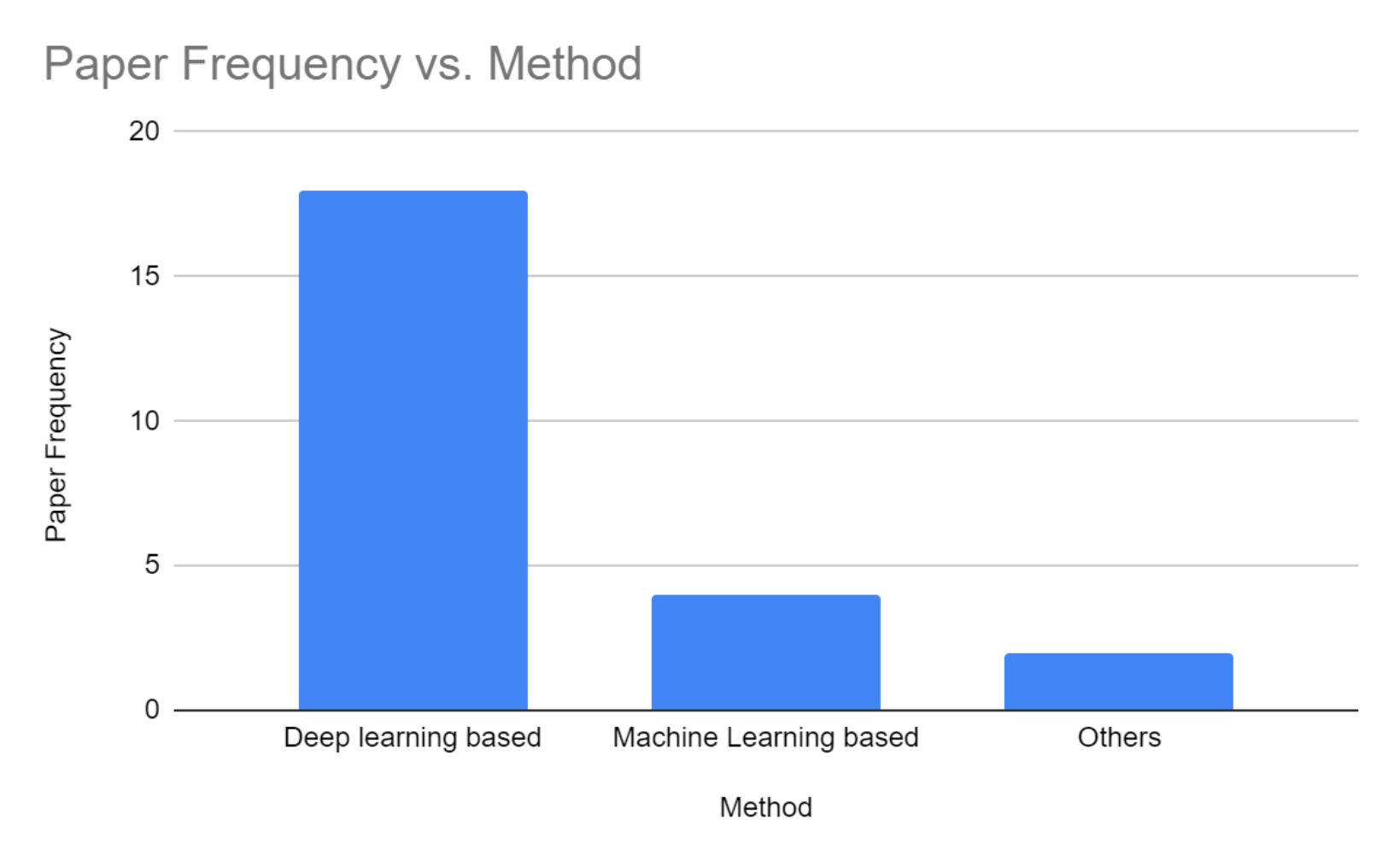}
\caption{Literature frequency of different AI techniques}
\label{fig:f6}
\end {center}
\end{figure}

COVID-19 has put researchers, health professionals at a critical situation due to the lack of timely information and historical data. Intelligent systems cannot work unless they are trained with reliable data. Application of AI and other related techniques: machine learning, deep learning is done based on the previous experiences i.e data and models. Given that very little information related to COVID-19 are available, researchers mostly rely on X-ray images and CT scan images. Although a  very small amount of Chest-X Ray (CXR) of COVID-19 is available, CXRs are prescribed as one of the first diagnostic tests by the physician. Most of the earlier research works use CXR images to detect COVID-19 (26.67\%) and others use chest CT images (Figure \ref{fig:f4}). The recent development of deep neural networks has opened up a new frontier in image classification. We found most of the research papers (43\%) use different architecture of deep neural networks (Figure \ref{fig:f6}) to classify images,  both CXR and CT scan (Table \ref{tab:table_details_deeplearning}). When it comes to image data, Convoluted Neural Networks dominated over all other algorithms and techniques(Table \ref{tab:table_algorithm}). Using CNN as a base, several studies come up with their architecture \cite{covidnet}. Our observation finds out among several CNNs, the Res-Net architecture as the most used one (Figure \ref{fig:f5}). Using the Res-Net architecture as a backbone, the models are different from one another on several parameters i.e. several hidden parameters, epochs and optimizer. Some studies also use a combination of different deep neural architectures. For example, Ezz et al.\cite{a39} compared six other different models with Red-NetV2 to propose the best one. Some studies, for example, Wang et al. \cite{a55} use a combination of Res-Net with other machine learning algorithms. The authors used Res-Net for feature extraction from CXR and leverage SVM and decision trees to develop a new algorithm. Other than classification, Biraja et al.\cite{a43} use different architectures i.e. bayesian CNN, monte Carlo drop weights along with Res-Net to assess the uncertainty associated with applying deep neural networks to detect COVID-19. Due to the pervasive use of deep learning networks and models, we have used a separate a table, Table \ref{tab:table_details_deeplearning} to present the deep learning architectures used and their results.

Other than deep learning approaches, traditional machine learning algorithms have also been applied when it comes to non-image data (Table \ref{tab:table_algorithm}). Using a combination of regression analysis and Group Method of Data Handling(GMDH), Pirouz et al. \cite{a46} tried to find a correlation and forecast based on demographic factors. Moreover, Yang et al.\cite{a45} combined epidemiological models with an ML model to show the effectiveness of the disease containment in China and predict the epidemic. With an addition to this, Loey et al.\cite{a36} used neural networks (GAN network) for detecting COVID-19, whereas Jiang et al.\cite{a42}  used decision tree, SVM based algorithm to detect COVID-19 from Xray images. 

The authors depend on the train-test split method for validation of their models as none of these models is used to test on real patients. As a validation metric, the studies have used accuracy, specificity, sensitivity, f-1 score and area under Receiver Operating Characteristic(ROC) curve (AUC). Among other evaluation matrics authors also used False Positive Rate (FPR), True Positive Rate (TPR), positive and negative predictive values (PPV and NPV respectively).  Our finding suggests that deep learning algorithms achieve a higher score in most of the evaluation matrices (see Table \ref{tab:table_algorithm}, and \ref{tab:table_details_deeplearning}).

\begin{singlespace}

\begin{table*}[htp]

\caption{A summary of the algorithm used in the literature for different objectives}
\small
\label{tab:table_algorithm}
\centering

    \begin{tabular}{|P{2cm}|P{3.5cm}| P{6.5cm}| P{4cm}| }
    \hline
    
    Objectives & 	Algorithms & 	Evaluation  Results & 	Literature\\
    \hline

   &  AI- based algorithm  
 	& -  & 
 	Rao et al.\cite{a60} \\
 	\cline{2-4}

 &	CNN & 
 	Accuracy (82.9\% ),
Specificity (80.5\%),
Sensitivity (84\%) & 	Wang et al. \cite{Wang}\\
\cline{2-4}

    & CNN   
 	&Accuracy (97.8\%)  &	Apostolopoulos et al. \cite{a29}
 	\\
 	\cline{2-4}

 	& CNN & 
 
 	F1-score (0.89)	 & Ezz et al. \cite{a39}
 	
 	\\
 	\cline{2-4}

 	& CNN &  
 	Sensitivity (100\%),
Specificity (100\%),
Accuracy (100\%),
F1- score (100\%)	& Salman et al. \cite{a56} 
	\\
 	\cline{2-4}

  Disease detection &	CNN &  
 	Accuracy (98\%),
Recall (96\%),
Specificity (100\%)	& Ali et al. \cite{a30}
	\\
 	\cline{2-4}

 &CNN,
SVM & 
 
 	Accuracy (98.27\%),
Sensitivity (98.93\%),
Specificity (97.60\%)
F-1 score (98.28\%),
Precision (97.63\%),    
Matthews Correlation Coefficient (96.54\%) & 
 
 	Umut et al.\cite{a31}
 	
 		\\
 	\cline{2-4}

 & CNN,SVM & 
 Accuracy(95.38\%), FPR(95.52\%), F1- score(91.41\%) , Kappa (90.76\%)& 
Sethy et al.\cite{sethy2020detection}
 		\\
 	\cline{2-4}
 	
 	 	& GAN Network & 
 	Accuracy (99.9\%)	&  Loey et al. \cite{a36}
 	 \\
 	\cline{2-4}
 	
 &	Decision trees, random forests and support vector machines & 
 	Accuracy(80\%) & 
 	Xiangao et al. \cite{a42}

 	 		\\
 	\cline{1-4}

Epidemic forecasting & 
 	Modified Auto-encoder for Modeling Time Series & 
 	  -	 & Hu et al. \cite{artificial}
 	  
 	  	 		\\
 	\cline{2-4}

	& Epidemiological model and ML based AI model & 
 	 -	& Yang et al. \cite{a45} 
 	  	
 	 		\\
 	\cline{1-4}

Sustainable development & 
 	Regression analysis and Group method of Data Handling & 
 	Accuracy (85.7\%) & 
 	Pirouz et al. \cite{a46}
 	
 	 		\\
 	\cline{1-4}

 & CNN and Grad Cam &
 	AUC  (0.989), 
Sensitivity (98.2\% ) &
 
 	Gozes et al.\cite{a37}
 	
 		  	 		\\
 	\cline{2-4}

Diseases diagnosis &CNN &
 	AUC (0.96)	&  Li et al.\cite{a38}
 	
 		  	 		\\
 	\cline{2-4}

 & XGBoost machine learning algorithm &
 
 	Death prediction accuracy (100\%),
Survival prediction accuracy (90\%)	& Li et al.\cite{a40}
	  	 		\\
 	\cline{2-4}
 
  & CNN &
 		-& 	Biraja et al. \cite{a43}
 		
 			  	 		\\
 	\cline{1-4}

    \end{tabular}

\end{table*}
\end{singlespace}

\begin{table*}[t]

\begin{singlespacing}

\caption{Summary of papers that used Deep Learning  in the field of COVID-19}
\small
\label{tab:table_details_deeplearning}
\centering

    \begin{tabular}{|p{2.1cm}|P{3cm}|P{3.8cm}| P{4.0cm}| P{3.2cm}| }
    \hline
    Literature  &	 Architecture	 & Task	 & Result & 	Research Outcome \\
    \hline

    Gozes et al. \cite{a27} & 
U-net, Resnet-50-2D &	Classification ,quantification and tracking: Covid-19 patients	& 0.996 (AUC)
98.2\% (Sensitivity)
92.2\%  (specificity)	& AI based software
\\
\hline

Wang et al \cite{a55} & 
ResNet-18	& Feature extraction from image data &	73.1\% (Accuracy)  67\% (specificity )and 
74\% (sensitivity) &	A CNN based algorithm leveraging decision tree and SVM
\\
\hline

Li et al. \cite{a38} & 
ResNet-50  &	Classification :
Covid-19 and pneumonia	 & 0.96(AUC)&	A CNN based model: COVnet
\\
\hline

Sethy et al. \cite{sethy2020detection} &
 Resnet50 &	Classification : 
COVID-19 & 95.38\%(Accuracy), 95.52\%(FPR), 91.41\%(F1- score) , 90.76\%(Kappa)&  	A CNN based model

\\

\hline

Apostolopoulos et al. \cite{a29} &
VGG19 ,
Mobile Net ,Inception ,Xception Inception ResNet v2  &	Classification: Covid-19,

Model Evaluation	& 97.82\% (Accuracy) &	A proposal: best deep learning network

\\
\hline

Ezz et al. \cite{a39} &
VGG19, DenseNet121, ResNetV2, InceptionV3, InceptionResNetV2, Xception, and MobileNetV2 &	Classification: Covid-19, 

Model Evaluation &	F1-scores : 
Normal :0.91
Covid-19 : 0.89	 & A proposal: best deep learning network

\\

\hline

Salman et al. \cite{a56} & 
InceptionV3 	& Classification: Covid-19 & 
	100\% (specificity) 100\% (accuracy) 100\% ( PPV)  100\%, (NPV)  100\% (f-1 score) &	A proposed model, implementation and evaluation
	
	\\
	
	\hline

	Ali et al. \cite{a30} &
ResNet50, InceptionV3 and Inception-ResNetV2 	& Classification: Covid-19&	98\% (accuracy) 96\% (recall) and 100\%(specificity) & 	A proposal: best deep learning network
\\

\hline

    \end{tabular}
\end{singlespacing}

\end{table*}

\section{FUTURE RESEARCH OPPORTUNITIES}

In this section, we have briefly presented the challenges and further research opportunities on AI/ML not only to fight against COVID-19 pandemic but also for the future pandemic. 

\subsection{Importance of study context of future research}
We observed that only one-third of the research (37\%) used contextual data while the rest (63\%) of them conducted research using data from more than one country. The countries or regions that are affected more than others have more opportunities to conduct contextual research. As China was the primary hotspot for the pandemic, a comparatively greater number of contextual studies have been conducted in China owing to the availability of more data and increased time to observe the nature of the pandemic. As the pandemic progressed, data from other countries also became available. Hence, there is considerably more scope for future contextual research that will aim to explore and predict the similarity of the pattern of the pandemic among Chinese studies and other regional studies.

\subsection{Potential areas of research}

The existing research has been conducted to detect and diagnose COVID-19, epidemic forecasting, sustainable development, and patient management. We observe that a relatively small number (11.4\%) of research has been conducted on epidemic forecasting, sustainable development and patient management. Further research can be done focusing on these areas. We observed that studies on epidemic forecasting and sustainable development used contextual data. We think that epidemic forecasting based research should always be contextual.  

\subsection{Collecting and storing various types of data}
There are opportunities to collect different types of data (e.g., images, texts, videos, etc.) and making it available for the researchers to conduct different experiments. Such efforts will be highly valuable for fighting the pandemic. 

\subsection{Disease diagnosis and prediction research using a large set of data}
Most disease detection and half of disease diagnosis-based research  were conducted using global data. However, we suggest more research in this direction could use diverse global data in the future for better performance. A few research studies (24\%) on disease detection and the other half research on disease diagnosis also performed context-based analysis on specific regions. Hence, future studies may consider other affected regions for disease pattern exploration as well. It can be insinuated from the studies considered in this review paper that a significant amount of data was not used for conducting machine learning and deep learning research. Future research could investigate if bigger datasets could result in better structured, authenticated and generalized outcomes. Additional work can be done to validate some of the original research studies (\cite{a28,a47}) that did not use data and only proposed models. The claims found in the studies can be explored more with sufficient data in the future.

\subsection{Exploring the effect and variation in research outcomes based on different types of data}

The majority of the existing research that has developed an AI/ML-based tool with the purpose of disease detection and progression has employed training and testing methods. The methods, in general, require related datasets to train and validate the systems to correctly predict the outcome of a given problem, which in this case is detecting the disease accurately. Several deep learning based research (48\%) studies have used chest X-Ray and CT scan images to determine the presence of opacity in lungs that depicts COVID-19. However, there are other lung diseases i.e., Pneumonia, COPD, Asthma etc. due to which similar effects on lungs X-Ray and CT images can be observed. Future studies can include other symptoms of patients in the format of text to train the system along with the images of lungs to determine COVID-19 from other diseases.

\subsection{Explore the contextual effect and variation in research outcomes}
Only one cross-country research has been conducted with the purpose of sustainable development, where correlation among confirmed cases, environmental and demographic factors of four different countries were calculated and compared. Other cross-country research studies, similar to the one stated above, could be pursued in the future to determine if the virus spread depends on the environmental factors. 

\subsection{Managing the ICU surge during the COVID-19 crisis}
It has been reported that the hospitals decided to provide service only to the young people, leaving elder citizens who have less survival possibility as the hospitals were out of capacity. Further research can be pursued to predict which patient has higher likelihood to be a critical case based on their medical history and symptoms. This would help the hospitals determine which patient can be cured at home and who will need ICU support. Research studies focusing on ICU admission would be helpful on early releasing of some patients, and thereby, leaving space for the patients who need it most. At the same time, studies can also prevent early termination from ICUs.  

\subsection{Supporting the health care workforce}
The lack of health professionals has been observed in highly affected areas. They had to work beyond their limit making them vulnerable to human errors. AI-assisted systems could be helpful here. A rule-based AI can monitor all the data in the ICU and suggests the professionals to take necessary steps. An efficient AI can help allocate and control the flow of oxygen level: a crucial treatment given to the COVID-19 patients. 

\subsection{Developing drugs/vaccine}
 Researchers all over the world currently have been working on developing drugs and vaccines for COVID-19. Different organizations have already been using AI to find a vaccine for COVID-19. From data analysis to decoy generation of COVID-19, AI can be helpful. Therefore, there are plenty of opportunities for improvement of these AI algorithms (Rosetta\cite{leaver2011rosetta3} and Quark\cite{xu2012ab}). Furthermore, AI can be used for simulation and analysis of different candidate vaccine.

 \section{CONCLUSION}

This paper presents a systematic review of exploring AI and ML techniques in fighting the COVID-19 pandemic. A total of 35 research articles were reviewed. These papers are analyzed and compared in various dimensions including the data types, input features, the AI techniques (the machine learning classification algorithms), as well as their objectives. Overall, there are three main contributions of our work. First, we have provided a summary of the findings in terms of objectives, data sources, data type, and volumes. Second, we have explored AI techniques that include various machine learning and deep neural network techniques in the field and compared their outcomes considering several popular evaluation metrics. Finally, we have identified several research issues based on our analysis and introduced corresponding new directions for future research. 

Our study has a number of limitations, but at the same time, it provides some avenues for future research. First, we have used some specific keywords for searching the relevant materials. Although, our search keywords provided effective results to achieve the goal of our study, there might be a risk to miss some important materials that did not emerge from our search queries. Second, we think timely and up-to-date materials related to coronavirus and AI techniques are the key things that we have identified, studied, and summarized in this paper. Therefore, future works would be needed to collect more resources. Third, in addition to exploring AI and ML techniques in fighting the Covid-19 pandemic, future research can be conducted to analyze data privacy and security in the relevant areas. Although we have several limitations mentioned above, our analysis and discussion can have significant implications for both the health practitioners and researchers. We believe that our review opens a promising path and can be used as a reference guide for future research in this area.

\balance
\bibliographystyle{IEEEtran}
\bibliography{citation}

\end{document}